\documentclass[11pt]{article}

\usepackage[final]{acl}

\usepackage{times}
\usepackage{latexsym}
\usepackage{xcolor}

\usepackage[T1]{fontenc}

\usepackage[utf8]{inputenc}

\usepackage{microtype}

\usepackage{inconsolata}

\usepackage{graphicx}
\usepackage{url}            
\usepackage{booktabs}       
\usepackage{amsfonts}       
\usepackage{nicefrac}       
\usepackage{colortbl}
\usepackage{makecell}
\usepackage{wrapfig}
\usepackage{amsmath}
\usepackage{amssymb}
\usepackage{tcolorbox}
\usepackage{algorithm}
\usepackage{tikz}
\usepackage{xspace}
\usepackage{array,multirow}
\usepackage{pgfplots}
\usepackage{verbatim}
\usepackage{caption}
\usepackage{bm}
\usepackage{booktabs}
\usepackage{framed}
\usepackage{stfloats}
\usepackage{iitem}
\usepackage{amssymb}
\usepackage{pifont}
\usepackage{arydshln}
\usepackage{enumitem}
\usepackage{algpseudocode}
\usepackage{threeparttable}
\usepackage{bbding}
\usepackage{wasysym}
\usepackage{changepage}
\usepackage{float}     
\usepackage{placeins}  
\usepackage{hyperref}       

\title{ConsistRM: Improving Generative Reward Models via Consistency-Aware Self-Training}

\author{
  \textbf{Yu Liang}\thanks{Equal contribution.}, \textbf{Liangxin Liu}\footnotemark[1], \textbf{Longzheng Wang}, \textbf{Yan Wang}, \textbf{Yueyang Zhang}\\ \textbf{Long Xia}, \textbf{Zhiyuan Sun}, \textbf{Daiting Shi}\thanks{ Corresponding authors.} \\
   Baidu Inc., Beijing, China \\
\texttt{\{liangyu05, liuliangxin, wanglongzheng, wangyan78\}@baidu.com} \\
\texttt{\{zhangyueyang, xialong01, sunzhiyuan01, shidaiting01\}@baidu.com}
}

\begin{document}
\maketitle
\begin{abstract}
Generative reward models (GRMs) have emerged as a promising approach for aligning Large Language Models (LLMs) with human preferences by offering greater representational capacity and flexibility than traditional scalar reward models. However, GRMs face two major challenges: reliance on costly human-annotated data restricts scalability, and self-training approaches often suffer from instability and vulnerability to reward hacking. To address these issues, we propose \textbf{ConsistRM}, a self-training framework that enables effective and stable GRM training without human annotations. ConsistRM incorporates the \textbf{Consistency-Aware Answer Reward}, which produces reliable pseudo-labels with temporal consistency, thereby providing more stable model optimization. Moreover, the \textbf{Consistency-Aware Critique Reward} is introduced to assess semantic consistency across multiple critiques and allocates fine-grained and differentiated rewards. Experiments on five benchmark datasets across four base models demonstrate that ConsistRM outperforms vanilla Reinforcement Fine-Tuning (RFT) by an average of 1.5\%. Further analysis shows that ConsistRM enhances output consistency and mitigates position bias caused by input order, highlighting the effectiveness of consistency-aware rewards in improving GRMs. Our implementation is available at \url{https://github.com/yuliangCarmelo/ConsistRM}.
\end{abstract}

\section{Introduction}
Large language models (LLMs) have seen growing popularity in both academia and industry due to their exceptional capability in tackling complex tasks. 
A core technique for aligning LLMs' outputs with human expectations is Reinforcement Learning from Human Feedback (RLHF)~\cite{ouyang2022training,bai2022training,dong2024rlhf,lambert2025reinforcement,lai2025survey}. 
The reward model in RLHF supplies supervision signals for training, and its effectiveness is widely acknowledged as a primary factor limiting the maximum achievable alignment performance\cite{rafailov2023direct,azar2024general}. 
Recently, an increasing number of studies have focused on generative reward models (GRMs)~\cite{mahan2024generative,ye2025improving,liu2025inference}. In contrast to traditional models that merely output scalar scores, GRMs offer greater representational capacity, enhanced generalization, and better adaptability to complex feedback.
 
\begin{figure}[t!]
    \centering
\includegraphics[width=1.0\linewidth]{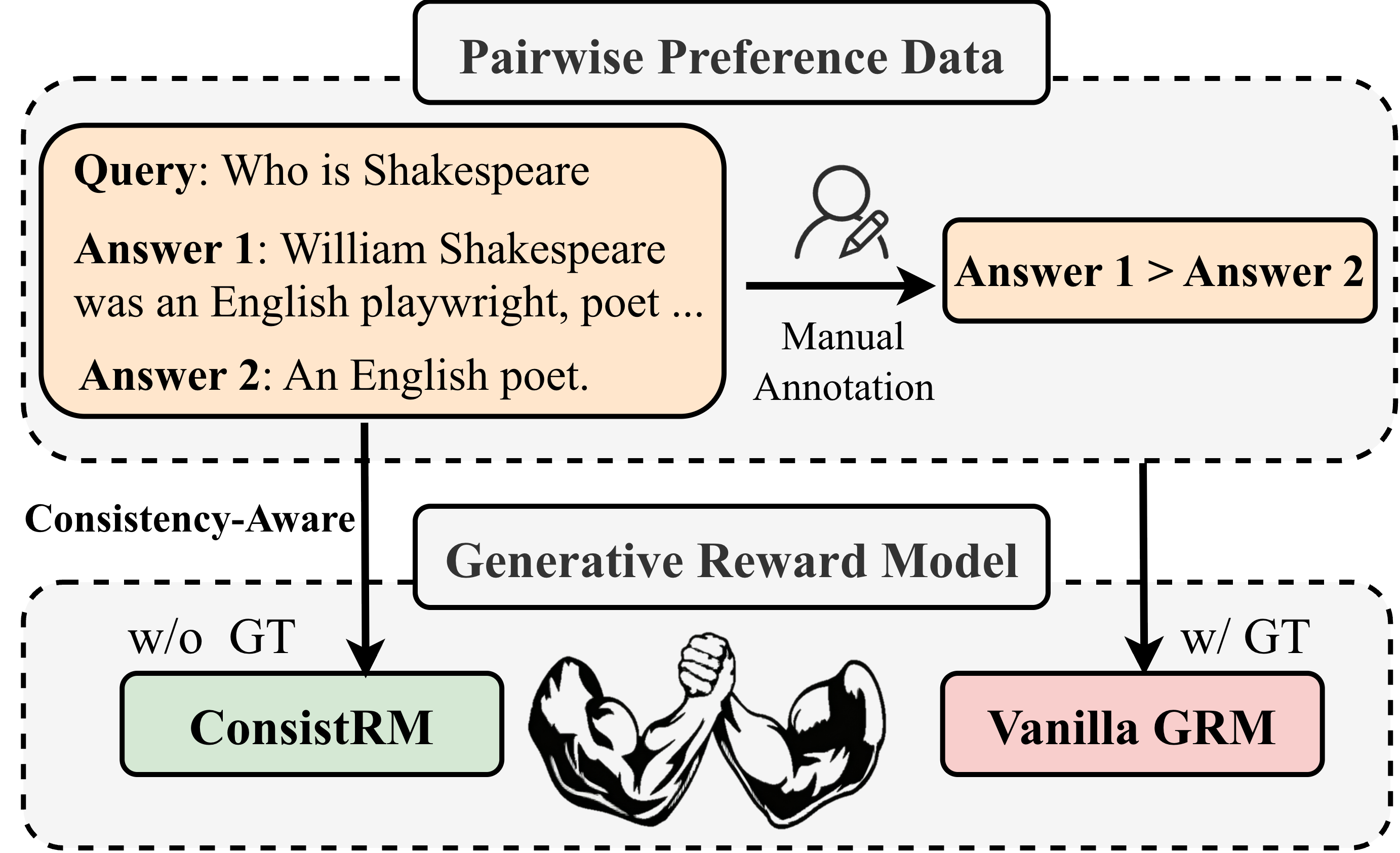}
    \caption{An overview of ConsistRM in comparison with vanilla GMRs, emphasizing its self-training framework based on consistency-aware learning to enhance model pairwise preference performance.}

    \label{example}
\end{figure}

While GRMs have demonstrated effectiveness, their training typically requires large amounts of high-quality human-annotated label data, making the process costly and hard to scale. For instance, methods like ~\cite{chen2025rm} depend on human-labeled reward signals for reinforcement learning. Similarly, approaches such as SynPref-40M~\cite{liu2025skywork} leverage LLMs to curate data automatically, though they still rely on initial human annotations for guidance. 
To reduce manual annotation, recent studies have explored unsupervised or self-training paradigms for GRMs, such as generating training signals from LLMs’ internal entropy or confidence ~\cite{zuo2025ttrl,shafayat2025can,wang2025survey}. 
However, these methods remain fragile because the reward signals are tightly correlated with the policy model, which easily leads to reward hacking and early overfitting to noisy pseudo-labels\cite{laidlaw2024correlated, zhang2025no}. Overall, despite notable advances, achieving robust and stable GRM training without any human feedback remains a major challenge.

%

In this work, we propose \textit{ConsistRM}, a novel self-training framework that leverages internally derived consistency signals without human-annotated labels. Our goal is to enhance the pairwise preference ability of GRM through two key modules:
(1) the Consistency-Aware Answer Reward, which considers temporal consistency (including both current model states and historical results) to provide reliable training signals;
(2) the Consistency-Aware Critique Reward, which measures semantic consistency across diverse critiques and offers fine-grained rewards.
The combination of these two complementary internal signals results in a more stable and reliable self-training framework.

We evaluate the effectiveness and generalization of ConsistRM across five benchmark datasets using four base models. Experimental results show that ConsistRM achieves an average performance gain of 1.5\% compared to vanilla Reinforcement Fine-Tuning (RFT). Through detailed analysis, we further demonstrate that ConsistRM enhances output consistency and mitigates position bias induced by input order.

Our contributions are summarized as follows:
\begin{itemize}[itemsep=1pt, topsep=1pt]

\item We propose ConsistRM, a framework that enables self-training for GRMs through consistency-aware mechanisms, eliminating the need for human feedback.

\item Within this framework, we design a method to construct pseudo-labels by leveraging consistency between current and historical model outputs, thereby providing GRMs with more reliable learning signals.

\item Empowered by the consistency-aware reward from ConsistRM, GRMs generate outputs with improved conciseness and accuracy, while significantly reducing position bias.

\end{itemize}

\section{Related work}
\paragraph{Generative Reward Model}

Recent experimental studies have demonstrated that GRMs represent a key new paradigm in reward modeling, primarily due to their enhanced interpretability and generalization capabilities~\cite{liu2025skywork,whitehouse2025j1,saha2025learning}. 
For example, DeepSeek‑GRM~\cite{liu2025inference} introduces a pointwise generative approach that produces detailed critiques and self-derived evaluation rules through reinforcement learning (RL), enabling more flexible and task-agnostic scoring compared to conventional scalar RMs. 
This generative paradigm has inspired subsequent research. For instance, RM‑R1~\cite{chen2025rm} incorporates chain‑of‑thought reasoning via a two‑stage process: distilling high‑quality reasoning traces followed by RL with verifiable rewards~\cite{guo2025deepseek}. 
This approach yields sample‑specific evaluation rationales, thereby enhancing both interpretability and empirical performance. Further extending the integration of reasoning, Reward Reasoning Models~\cite{guo2025reward} introduce a “reason-before-judgment” methodology without requiring annotated traces and dynamically adapt computation based on input complexity.
However, a common limitation of these existing generative and reasoning‑based GRMs is their dependence on supervised signals and multi‑stage pipelines to obtain high‑quality labels or reasoning traces, which substantially limits their further development.

\paragraph{Label-Free Self-Training for LLMs}
To address the challenge of acquiring high-quality labels, researchers have shifted their focus toward self-training paradigms that leverage LLMs to generate supervisory signals, thereby reducing reliance on human annotation. Existing methods can be divided into two main categories:
The first focuses on stabilizing outputs by encouraging low-entropy predictions or leveraging internal confidence signals to improve training efficiency ~\cite{li2025generalist,zhang2025consistent,li2025jointly}. The second constructs supervision signals from the consistency of multiple output paths, providing more reliable guidance for reward modeling, which is closer to our approach. A representative example of the second category is Test-Time Reinforcement Learning (TTRL), which uses majority voting over candidate answers as pseudo-labels for RL updates ~\cite{zuo2025ttrl,liu2025ettrl}. Extensions such as EVOL-RL~\cite{zhou2025evolving} and collaborative reward frameworks~\cite{zhang2025co} have been proposed to enhance this paradigm. However, recent studies~\cite{wang2025beyond} suggest that pure majority voting may suffer from spurious consensus when candidate outputs are highly correlated, offering limited guidance for fine-grained reasoning. Other approaches exploit internal model states, using confidence scores for advantage estimation or as direct feedback in preference optimization~\cite{zhao2505learning,van2025post}. Nevertheless, methods relying solely on within-round consensus or coarse-grained feedback are highly noise-sensitive, compromising the reliability of the resulting supervisory signals.

\begin{figure*}[!t]
    \centering
    \includegraphics[width=0.95\textwidth]{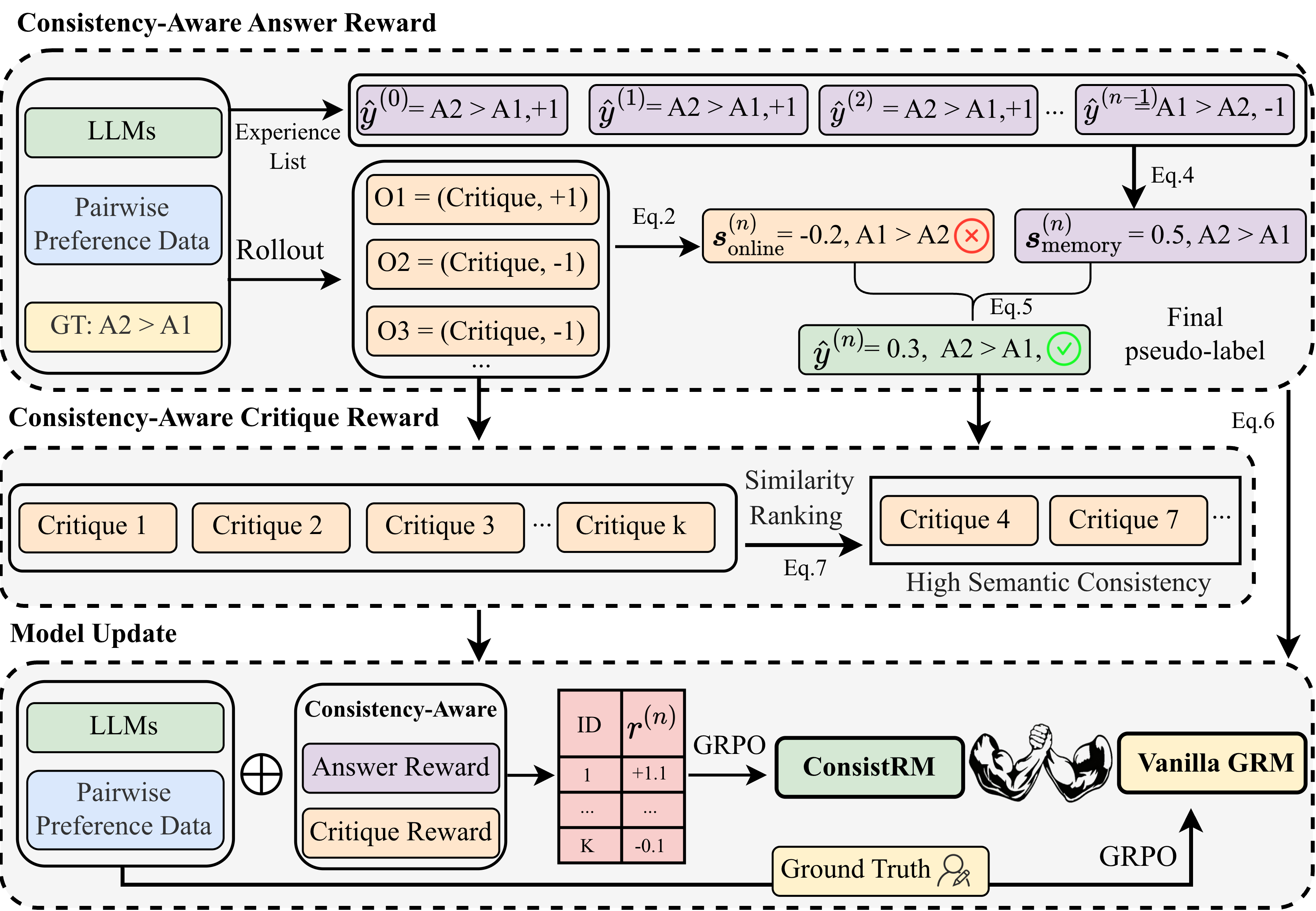} 
    \caption{Overall Framework of ConsistRM.
    The framework aims to produce an output $o = (c, y)$, which comprises a textual critique $c$ and a binary preference label $y$. Here, $y = -1$ indicates that  $a_1$ is preferred over $a_2$ ($a_1 \succ a_2$)., and $y = 1$ indicates the opposite preference ($a_2 \succ a_1$).
    Consistency-Aware Answer Reward module integrates online-state preference and memory-driven preference to generate pseudo-labels for training. 
    The Consistency-Aware Critique Reward module provides fine-grained rewards by measuring the similarity of multiple criteria. 
    By combining these two rewards, ConsistRM performs self-training to progressively enhance the GRMs’ ability.}
    \label{figure:main_pipeline}
\end{figure*}


\section{Method}
We propose ConsistRM, a stable and reliable self-training framework that explicitly GRM both answer preference consistency and critique consistency. As illustrated in Figure 2, ConsistRM consists of two core components: (1) Consistency-Aware Answer Reward(CAAR), which considers online-state and memory-driven preference consistencies to produce reliable pseudo-labels; (2) Consistency-Aware Critique Reward(CACR), which encourages semantic consistency among multiple generated critiques, thereby improving the stability of the analysis process. ConsistRM leverages consistency between answers and critiques to provide fine-grained reward, thereby enhancing its pairwise preferences ability.

\subsection{Pairwise Preference Formulation}
We model the problem of pairwise preference comparison as a conditional generative process. Given an input $x = (q, a_1, a_2)$ consisting of a query $q$ and two candidate responses $(a_1, a_2)$, GRM $\pi$ generates a structured output $o = (c, y)$, where $c$ is a textual critique and $y$ is a binary preference label, as detailed below:
\begin{equation}
(c, y) \sim \pi_{\theta}^{(n)}(c, y \mid q, a_1, a_2),
\end{equation}
where ${(n)}$ denotes the GRM training at the $n$-th iteration. The preference label $y \in  \{-1, 1\}$ indicates the direction of preference: $y = -1$ corresponds to $a_1 \succ a_2$, while $y = 1$ corresponds to $a_2 \succ a_1$.

\subsection{Consistency-Aware Answer Reward}
To alleviate reward hacking\cite{gao2023scaling, laidlaw2024correlated} and establish more reliable pseudo labels, we propose a Consistency-Aware Answer Reward (CAAR) mechanism. This mechanism constructs pseudo-labels by building the relation between two types of preferences: the Online-State Preference (derived from the current training iteration) and the Memory-Driven Preference (captured from memory experience). 
Consequently, the pseudo-labels produced are more robust and reliable, providing stronger support for self-training.

\paragraph{Online-State Consistency Preference}
We define the online-state preference signal as a metric that quantifies preference consistency under the current state of GRMs. Given an input $x = (q, a_{1}, a_{2})$, the model performs $K$ rollouts during training, generating a set of outputs:
$O = [o_1, o_2, \dots, o_k]$, and $o_j = (c_j, y_j)$ corresponds to the output of model from the $j$-th rollout.
To quantify the online-state preference consistency for the input $x$ at $n$-th iteration, we aggregate the preference prediction $y_j$ obtained from all rollouts in $O$ and the online-state consistency preference  $s_{\text{online}}^{(n)}$ is defined as:
\begin{equation}
\boldsymbol{s}_{\text{online}}^{(n)} = \frac{1}{K} \sum_{j=1}^{K} {y}_{j}
\end{equation}
where $n$ means the $n$-th training iteration.

\paragraph{Memory-Driven Consistency Preference.}
Although online-state preference consistency reflects the preferences expressed by the current GRM $\pi$, the estimate can be unreliable in the early stages of training, which may cause the GRM to converge toward suboptimal local optima\cite{zhang2025no}. To enhance the robustness and reliability of pseudo-labels, we propose a memory-driven consistency preference.
Specifically, for each input sample $x$, we maintain a dynamically updated experience list $\mathcal{E}$ that stores the pseudo-labels collected across training iterations:
\begin{equation}
\mathcal{E}^{(n-1)} = [\hat{y}^{(0)}, \hat{y}^{(1)}, \ldots, \hat{y}^{(n-1)}]
\end{equation}
where $\hat{y}^{(n-1)}$ denotes the pseudo-label constructed for input $x$ at the ($n-1$)-th training iteration. Notably, the initialization pseudo-label $\hat{y}^{(0)}$ is generated by the initial GRM $\pi_{\theta}^{(0)}$.
To quantify the memory-driven consistency preference for input $x$, we aggregate all pseudo-labels accumulated in $\mathcal{E}$ by computing their average. This simple averaging strategy serves as a stable semantic anchor over historical predictions, rather than emphasizing any individual pseudo-label. At the $n$-th training iteration, the memory-driven preference consistency ${s}_{\text{memory}}^{(n)}$ is defined as:
\begin{equation}
\boldsymbol{s}_{\text{memory}}^{(n)} = \frac{1}{n - 1} \sum_{i=0}^{n-1} \hat{y}^{(i)}
\end{equation}

\paragraph{Pseudo-Label with Temporal Consistency}
We derive the final pseudo-label $\hat{y}$ for the current iteration by jointly leveraging consistency signals at multiple levels, which leads to more reliable pseudo-labels that are better aligned with the training dynamics. Specifically, for each input sample $x$ at the $n$-th training iteration, the pseudo-label is constructed as follows:
\begin{equation}
{\hat{y}}^{(n)} = sgn(s_{\text{online}}^{(n)} + {s}^{(n)}_{\text{memory}})
\end{equation}
where $\operatorname{sgn}(\cdot)$ denotes the \textit{sign function}. This ternary labeling strategy explicitly assigns $\hat{y} = 0$ in the presence of inconsistent preference signals, which prevent low-confidence supervision from dominating the optimization. Although early-stage pseudo-labels may be noisy, their influence naturally diminishes as more iterations are accumulated, making the overall estimate increasingly robust over time, as further validated by our analysis of CAAR in Appendix~\ref{Effectiveness of CAAR}.

Finally, for each input $x$, we define the Consistency-Aware Answer Reward $r_j^{(a)}$, which measures the agreement between the constructed pseudo-label $\hat{y}^{(n)}$ and the preference prediction $y_j^{(n)} $ generated in the $j$-th rollout at $n$-th training iteration. The reward is computed as:
\begin{equation}
\begin{split}
r_j^{(a, n)} &= \mathbb{I}\!\left[\hat{y}^{(n)} = y_j^{(n)}\right] \\
&\quad - \mathbb{I}\!\left[\hat{y}^{(n)} \neq y_j^{(n)} \;\wedge\; \hat{y}^{(n)} \neq 0\right]
\end{split}
\end{equation}
where $\mathbb{I}[\cdot]$ denotes the \textit{indicator function}, and $r_{j}^{(a,n)} \in \{-1, 0, 1\}$.

\subsection{Consistency-Aware Critique Reward}
Rather than relying solely on outcome-level supervision, we introduce a Consistency-Aware Critique Reward (CACR) in ConsistRM to deliver richer reward signals and stabilize the outputs of GRMs.
We posit that high semantic similarity among multiple generated critiques implies that their corresponding responses lie within a similar high-reward semantic region, which can be leveraged to guide model optimization more effectively.
When the GRM $\pi_\theta$ produces critiques with strong semantic consistency for the same input $x$, the resulting reward signal becomes more stable and less sensitive to sampling variability. 
Based on this observation, we use consistency among critiques as a proxy for reward signal reliability, and that reinforcing such consistency improves model alignment with stable and representative evaluations.

Concretely, following \cite{zhou2025evolving}, each critique $c_{j}^{(n)}$ is encoded into a dense vector using the Qwen3-4B-Embedding \cite{qwen3embedding} and computes the cosine similarity matrix $S^{(n)}$. Based on this matrix, we rank the critiques $c_{j}^{(n)}$ according to their semantic consistency, and select the top-$p$ critiques to receive the reward. The corresponding reward is defined as:
\begin{equation}
r_{j}^{(c, n)} =
\begin{cases}
0.1, & \text{if } y_{j}^{(n)} = \hat{y}^{(n)} \text{ and } \mathrm{rank}(c_{j}^{(n)}) \le p; \\
0,   & \text{otherwise.}
\end{cases}
\end{equation}
where $r_{j}^{(c, n)}$ represents the consistency-aware critique reward for input $x$.

\subsection{Final Reward}

We define the final reward for each generated response by integrating three components: a mandatory format constraint, a consistency-aware answer reward, and a consistency-aware critique reward. 
The format constraint is first applied to enforce structural validity. The correctness and stability reward components are computed only if this constraint is satisfied.
Thus, for an input $x$ at the $j$-th rollout of the $n$-th training iteration, the final reward is computed as follows:
\begin{equation}
r^{(n)} =
\begin{cases}
-5, & \text{if invalid}; \\[6pt]
r_{j}^{(a,n)} + r_{j}^{(c,n)}, & \text{if } \hat{y}^{(n)} \neq 0; \\[4pt]
0, & \text{if } \hat{y}^{(n)} = 0.
\end{cases}
\end{equation}
Details of the ConsistRM algorithm are provided in the appendix~\ref{Algorithm}.





\section{Experiments}
\label{data selection exp}
\subsection{Setup}

\paragraph{Benchmark}
To thoroughly assess ConsistRM's performance on pairwise preference tasks, we utilize five reward model benchmarks that contain multilingual instructions and responses from diverse LLMs:
(1) \textbf{RewardBench}~\cite{lambert2024rewardbench} includes 2,985 evaluation samples from 23 data sources, categorized into four main groups: chat, chat-hard, safety, and reasoning.
(2) \textbf{Preference Proxy Evaluations (PPE)}~\cite{frick2025how}: we focus on its Preference subset with Chatbot Arena human preference pairs, covering 20 LLMs across 121+ languages.
(3) \textbf{RM-Bench}~\cite{liu2025rmbench}: a benchmark designed to assess reward model robustness through sensitivity to content differences and style biases.
(4) \textbf{RMB}~\cite{zhou2024rmb} is a comprehensive benchmark with 49 real-world task categories focused on helpfulness and harmlessness. It uses synthetically generated preference pairs and GPT-4 for pointwise ratings based on query-specific principles. Human verification ensures dataset quality.
(5) \textbf{JudgeBench}~\cite{tan2025judgebench}: we report results for the subset generated by GPT-4o and use position-consistent accuracy, considering a sample correct only if the judge gives the right verdict in both response orders.

\paragraph{Implementation details}
We use Qwen3 as our testbed and fine-tune it for 4 epochs using the GRPO~\cite{shao2024deepseekmath}. The fine-tuning employs a global batch size of 64 and a learning rate of 1e-6, with a maximum generation length of 1024 tokens. GRPO involves a proximal policy optimization (PPO) phase; during this phase, we use a mini-batch size of 32 and perform 8 rollouts per update step. The generation temperature during training is set to 1.0. To stabilize training and prevent excessive policy drift, we apply KL regularization with a coefficient of 0.001.
For inference, following ~\cite{whitehouse2025j1,saha2025learning}, we set the maximum generation length to 2048 tokens and the temperature to 0 (more detailed configuration can be found in Appendix \ref{A1}).

\paragraph{Baselines} 

We perform the following experiments to evaluate our approach comprehensively:
\begin{itemize}[leftmargin=1em]
\item \textbf{Base}: We evaluate the Qwen3-4B~\cite{yang2025qwen3} and Qwen3-8B models on benchmark datasets without any fine-tuning.
\item \textbf{SFT}: We fine-tune the base Qwen3 models using supervised fine-tuning on the HelpSteer3 dataset~\cite{wang2025helpsteer3}.
\item \textbf{RFT}: we further fine-tune the Qwen3 models using the GRPO algorithm~\cite{shao2024deepseekmath} for reinforcement fine-tuning (for more dataset detailed can be found in Appendix \ref{A2}).

\item \textbf{TTRL}: This method uses majority voting over candidate answers as pseudo-labels for reinforcement learning updates~\cite{zuo2025ttrl,liu2025ettrl}.

\end{itemize}

We then compare our ConsistRM against several open-source GRMs of similar scale, including EvalPlanner-Llama-8B~\cite{saha2025learning}, J1-Llama-8B~\cite{whitehouse2025j1}, and DeepSeek-GRM-27B~\cite{liu2025inference}

\begin{table*}[t]
\centering
\scalebox{0.75}{
\begin{tabular}{l c ccccc  ccc  }
\toprule

\multirow{2}{*}{\bf ID ~ System} & \multirow{2}{*}{\textbf{GT}} & \multirow{2}{*}{\textbf{RewardBench}} & \multirow{2}{*}{\textbf{PPE Pref}}&  \multirow{2}{*}{\textbf{RM-Bench}}& \multirow{2}{*}{\textbf{RMB}} &  \multirow{2}{*}{\textbf{JudgeBench }} & \multicolumn{2}{c}{\textbf{Overall}}  \\

\cmidrule(lr){8-9}

&    &  &    &  &  &   & \text{{AVG}} & \textit{$\Delta$ ($\uparrow$)}\\

\midrule &\multicolumn{5}{c}{\it Open Generative Reward Models} \\ 
1 ~~~~~EvalPlanner-Llama-8B &  \textcolor{green}{\Checkmark} &83.0 & 54.3 & 68.1 & - &  30.2 &- &-  \\

2 ~~~~~J1-Llama-8B  & \textcolor{green}{\Checkmark}   &85.7 & 59.8 & 73.4 & - &  42.0 &- &-  \\

3 ~~~~~DeepSeek-GRM-27B & \textcolor{green}{\Checkmark}   & 88.5 & 67.2 & - & 70.3 &  - &- &-  \\
\midrule &\multicolumn{5}{c}{\it Implemented Existing  Method} \\ 

4 ~~~~~Qwen3-4B &  &78.6 & 63.3 & 74.5 & 76.4 & 50.0 & 68.6& -    \\

5 ~~~~~4 + SFT & \textcolor{green}{\Checkmark} &79.8 & 63.1 & 76.1 & 76.5 & 43.7 & 67.8 &  \textcolor[RGB]{255,25,0}{-0.8}   \\

6 ~~~~~4 + RFT  & \textcolor{green}{\Checkmark}   & 80.0 &\underline{64.7}& 76.6& 76.7 & 51.1 & 69.8  &  \textcolor[RGB]{0,176,80}{+1.2}  \\

7 ~~~~~4 + TTRL  &  \textcolor{red}{\XSolidBrush}  &\textbf{81.6} & 61.8 & 76.3 & 75.6 & \underline{52.3} & 69.5 &  \textcolor[RGB]{0,176,80}{{+0.9}}  \\
\hdashline 

\multicolumn{10}{c}{\it Our Method} \\

8 ~~~~~4 + CAAR  &  \textcolor{red}{\XSolidBrush}  & \underline{81.4} & 64.1 & \underline{76.6} & \textbf{77.5} & 52.0 & 70.3 & \textcolor[RGB]{0,176,80}{{+1.7}}  \\
\rowcolor[HTML]{C0C0C0} 

9 ~~~~~8 + CACR (ConsistRM)  &  \textcolor{red}{\XSolidBrush} & 80.3 & \textbf{65.1} & \textbf{76.8 }& \underline{76.5} & \textbf{56.1} & 71.0  & \textcolor[RGB]{0,176,80}{\textbf{+2.4}}  \\

\midrule &\multicolumn{5}{c}{\it Implemented Existing  Method} \\ 
10 ~~~~Qwen3-8B &   &  81.6 & 63.8 & 75.8 & \underline{78.8} & 54.3 & 70.9 &-  \\

11 ~~~~10 + SFT & \textcolor{green}{\Checkmark} & 82.7 & 65.0 & 77.1 & 76.9 & 51.7 & 70.7  & \textcolor[RGB]{255,25,0}{-0.2}   \\

12 ~~~~10 + RFT  & \textcolor{green}{\Checkmark}   & \underline{85.4} & \underline{65.4} & \underline{78.2} & {78.2} & 55.4 & 72.5  &  \textcolor[RGB]{0,176,80}{+1.6}  \\

13 ~~~~10 + TTRL  &  \textcolor{red}{\XSolidBrush} & 85.3 & 65.0 & 77.4 & 74.2 & \underline{56.8} & 71.7 &    \textcolor[RGB]{0,176,80}{+0.8}\\

\hdashline 
\multicolumn{10}{c}{\it Our Method} \\

14 ~~~~10 +  CAAR   &  \textcolor{red}{\XSolidBrush}  &84.9 & 64.8 & 77.3 & 78.1 & 56.0 & 72.2  & \textcolor[RGB]{0,176,80}{{+1.3}}  \\
\rowcolor[HTML]{C0C0C0} 
15 ~~~~14 + CACR (ConsistRM)  &  \textcolor{red}{\XSolidBrush}  & \textbf{85.6} & \textbf{67.7} & \textbf{78.3} &\textbf{79.1} & \textbf{56.9} & 73.5
 & \textcolor[RGB]{0,176,80}{\textbf{+2.6}}  \\

\bottomrule
\end{tabular}
}

\caption{{Overall performance of different GRMs on five benchmarks. ConsistRM achieved the best results without requiring any human-annotated training data, demonstrating our effectiveness of the self-training paradigm.
 \textbf{Bold} indicating the best performance and \underline{underline} indicating the second-best performance.}
}
\label{table:main LLMs}
\end{table*}

\subsection{Main Results}
We focus our analysis on Qwen3-8B, which is representative of the overall trends observed across smaller 4B models. System (10) shows the original Qwen3-8B’s performance on pairwise preference benchmarks. Systems (5) and (11) achieve only limited improvements through vanilla SFT. The core issue is that standard loss functions (e.g., cross-entropy) average over the entire dataset, smoothing out training signals and impeding targeted updates for hard samples.
However, online reinforcement learning offers an effective solution to this challenge. Specifically, GRPO assigns differentiated rewards to multiple sampled outputs based on existing labels, thereby yielding stable performance improvements, as shown in Systems~(6) and~(12).
In contrast, simple unsupervised learning results in marginal improvements, as shown in Systems~(7) and~(13), primarily due to the increasing bias in pseudo-labels during the later stages of training.

Additionally, Systems (14) shows that incrementally integrating the ConsistRM submodule enhances GRM performance, surpassing the contemporary vanilla RFT and TTRL method. This stems from stable and reliable learning labels derived from current and historical state decisions, enabling effective self-learning.
Notably, ConsistRM significantly boosts Qwen3’s pairwise preference performance compared to vanilla RFT and other advanced methods, as seen in System (15). 
The improvement is particularly pronounced on JudgeBench benchmarks, likely due to our ConsistRM produce more consistent responses.

\subsection{Ablation Study}
To better understand the contribution of each component in ConsistRM, we conduct ablation studies by removing individual modules. As shown in  Table~\ref{tab:ablation_consistrm}, removing either the CACR or the Online-State Consistency Preference module leads to an average performance drop of approximately 1.2 points, indicating that both components play essential roles in the overall framework. In contrast, ablating the Memory-Driven Consistency Preference results in the most significant degradation (-3.2), suggesting that this component is critical for maintaining high-quality pseudo-labels. Its removal substantially reduces pseudo-label accuracy, indicating that accurate historical aggregation is crucial for stabilizing pseudo-label quality and improving performance.
\begin{table*}[t]
\centering
\small
\setlength{\tabcolsep}{4pt}
\renewcommand{\arraystretch}{1.1}
\begin{tabular}{lccccccc}
\toprule
System & RewardBench & PPE Pref & RM-Bench & RMB & JudgeBench & Avg &\textit{$\Delta$ ($\uparrow$)}\\
\midrule
ConsistRM & 85.6 & 67.7 & 78.3 & 79.1 & 56.9 & 73.5 & -\\
\quad w/o CACR & 84.9 & 64.8 & 77.3 & 78.1 & 56.0 & 72.2 & \textcolor[RGB]{255,25,0}{-1.3}\\
\quad w/o Online-State Consistency Preference & 85.5 & 64.1 & 78.6 & 76.7 & 56.7 & 72.3 & \textcolor[RGB]{255,25,0}{-1.2}\\
\quad w/o Memory-Driven Consistency Preference & 84.3 & 63.1 & 75.4 & 74.2 & 54.8 & 70.4 & \textcolor[RGB]{255,25,0}{-3.2}\\
\bottomrule
\end{tabular}
\caption{Ablation study results of ConsistRM on five benchmarks.}
\label{tab:ablation_consistrm}
\end{table*}

\section{Analysis}


This section aims to address the following research questions through experiments:
(1) Can ConsistRM adapt to different foundation models or other rewards of critique? (see §\ref{Ablation Study1 of ConsistRM})
(2) What are the advantages of ConsistRM? (see §\ref{Ablation Study2 of ConsistRM})
(3) How efficient is ConsistRM? (see §\ref{Ablation Study3 of ConsistRM})
For evaluation, we primarily use Qwen3-4B as the base model and test our approach on benchmarks: RewardBench and JudgeBench, unless otherwise specified.

\subsection{The Varieties of ConsistRM}
\label{Ablation Study1 of ConsistRM}
\paragraph{Various Foundation Models}
Although ConsistRM shows impressive improvements on Qwen3-4B and 8B, its generalization to other foundation models remains to be validated. To address this, we apply ConsistRM to different models, including Qwen3-14B and LLaMA-3.1-8B. Under consistent experimental settings, we evaluate performance on two standard benchmarks. As shown in Table \ref{tabel_various_model}, ConsistRM consistently improves both models across these benchmarks (+1.3, +1.6). 
These results demonstrate that ConsistRM can effectively leverage the inherent consistency ability of LLMs (of varying sizes and architectures) for effective self-training without any human feedback.

\begin{table*}[t]
\centering
\scalebox{0.72}{
\begin{tabular}{l c c c c c c c c c}
\toprule
\multirow{2}{*}{\bf Foundation Model} &\multirow{2}{*}{\textbf{Method}} & \multirow{2}{*}{\textbf{RewardBench}} & \multirow{2}{*}{\textbf{PPE Pref}} & \multirow{2}{*}{\textbf{RM-Bench}} & \multirow{2}{*}{\textbf{RMB}} & \multirow{2}{*}{\textbf{JudgeBench}} &   \multicolumn{2}{c}{\textbf{Overall}}  \\
\cmidrule(lr){8-9}

&    &  & & & & & \text{{AVG}} & \textit{$\Delta$ ($\uparrow$)}\\
\midrule
\multirow{2}{*}{\bf Qwen3-4B}& Vanilla RFT & 80.0 &64.7& 76.6& 76.7 & 51.1 & 69.8  & -  \\
& ConsistRM & 80.3 & 65.1 & 76.8 & 76.5 & 56.1 & 71.0  & \textcolor[RGB]{0,176,80}{\textbf{+1.2}}  \\
\midrule 
\multirow{2}{*}{\bf Qwen3-8B}& Vanilla RFT  & 85.4 & 65.4 & 78.2 & 78.2 & 55.4 & 72.5  &  -  \\

& ConsistRM &85.6 & 67.7 & 79.2 & 79.2 & 55.9 & 73.5  & \textcolor[RGB]{0,176,80}{\textbf{+1.0}}  \\
\midrule
\multirow{2}{*}{\bf Qwen3-14B}& Vanilla RFT& 87.4 & 66.3 & 79.3 & 76.3 & 68.6 & 75.6 & -   \\

& ConsistRM  & 89.0 & 67.7 & 77.5 & 78.4 & 71.7 & 76.9 &\textcolor[RGB]{0,176,80}{\textbf{+1.3}}  \\
\midrule
\multirow{2}{*}{\bf Llama-3.1-8B}& Vanilla RFT & 74.8 & 55.3 & 63.2  & 62.7 & 30.2 & 57.2   \\

& ConsistRM  &76.6 & 54.3 & 64.8 & 64.9 & 33.4 & 58.8&  \textcolor[RGB]{0,176,80}{\textbf{+1.6}}  \\
\bottomrule
\end{tabular}}
\caption{Results of ConsistRM with various foundation models. ConsistRM can be stably applied to LLMs of different sizes and architectures.}
\label{tabel_various_model}
\end{table*}

\paragraph{Various Critique Rweard}
This section proposes two modifications to the Consistency-Aware Critique Reward Module (CACR) to examine its rationality.
In contrast to ConsistRM, we introduce an additional reward term for samples in the bottom 50\% of semantic consistency, denoted as CACR-Low. Second, we replace the critique quality estimation in CACR with a token-level confidence measure—DeepConfidence ~\cite{fu2025deep}, which is implemented in two variants:
(1) DeepConf-Tail: the average confidence of the final 128 tokens, based on the observation that reasoning quality often wanes at the end of long chains, where final steps are crucial for correctness.
(2) DeepConf-Low: the average confidence of the 128 lowest-confidence tokens across the entire reasoning trajectory.

The effectiveness and necessity of CACR are supported in the Table~\ref{tabel_quality_Trajectory}. First, performance deteriorates when high rewards are assigned to low-similarity critiques, suggesting that higher-similarity critiques reflect stronger consensus and thus support more reliable preference judgments.
Second, simply replacing CACR with DeepConfidence causes a significant performance drop (from 68.2 to 65.4). 
We attribute this phenomenon to reward hacking, in which optimization over token-level confidence biases the model toward overly confident but erroneous tokens, thereby leading to homogenized outputs and degraded overall performance.
The above experiments show the rationality of our proposed consistency-aware critique reward.

\begin{table}[t]
\centering
\scalebox{0.64}{
\begin{tabular}{l c c c c}
\toprule
\multirow{2}{*}{\bf System} & \multirow{2}{*}{\textbf{RewardBench}} & \multirow{2}{*}{\textbf{JudgeBench}} &  \multicolumn{2}{c}{\textbf{Overall}}  \\
\cmidrule(lr){4-5}
&    &  & \text{{AVG}} & \textit{$\Delta$ ($\uparrow$)}\\
\midrule
\multicolumn{5}{c}{\it Various  Critique Rweard} \\ 
ConsistRM &80.3 & 56.1 & 68.2& -\\
\quad {w/ CACR-Low} & 81.2 & 50.6& 65.8  &   \textcolor[RGB]{255,25,0}{-2.4}      \\
\quad {w/ DeepConf-Tail} &80.4 & 50.3 & 65.4 & \textcolor[RGB]{255,25,0}{-2.8}   \\
\quad {w/ DeepConf-Low} &79.8 & 51.2 & 65.5 & \textcolor[RGB]{255,25,0}{-2.7}  \\
\midrule
\multicolumn{5}{c}{\it Positional Bias} \\ 
Qwen3-4B & 69.7 & 50.0 & 59.9 & -  \\
Vanilla RFT & 71.4 & 51.1 &61.3 &\textcolor[RGB]{0,176,80}{{+1.4}}  \\
ConsistRM &74.2 & 56.1 & 65.2 &\textcolor[RGB]{0,176,80}{{+5.3}}   \\
\bottomrule
\end{tabular}}
\caption{Comparison results with different setups on GRMs. ConsistRM achieves the better Results.}
\label{tabel_quality_Trajectory}
\end{table}

\subsection{The Consistency of ConsistRM}
\label{Ablation Study2 of ConsistRM}





\paragraph{Robustness to Positional Bias} 
In this section, we evaluate how effectively ConsistRM mitigates position bias in GRMs.
Position bias is defined as the tendency of a model’s preference between two responses to depend on their presentation order. In practice, for the same query, swapping the response order can lead to a reversal of the model’s preference.
To quantify this bias, we adopt the evaluation standard from ~\citet{tan2025judgebench}, where a sample is considered correctly evaluated only if the model produces consistent judgments under both orderings of the answer pair. We apply this standard to both RewardBench and JudgeBench, comparing base models, RFT models, and our ConsistRM.

As shown in Table \ref{tabel_quality_Trajectory}, vanilla RFT yields only limited improvement in mitigating positional bias (+1.4). We hypothesize that for challenging samples, the model tends to learn shallow prompt–answer mappings rather than robust reasoning. 
In contrast, ConsistRM leverages its ability to generate more suitable training labels, thereby promoting the training process and substantially mitigating positional bias. This leads to an improvement in the consistency score from +1.4 to +5.3, showing the robustness of ConsistRM.

\begin{figure}[t]
	\centering
 \scalebox{0.22}{
	\includegraphics[]{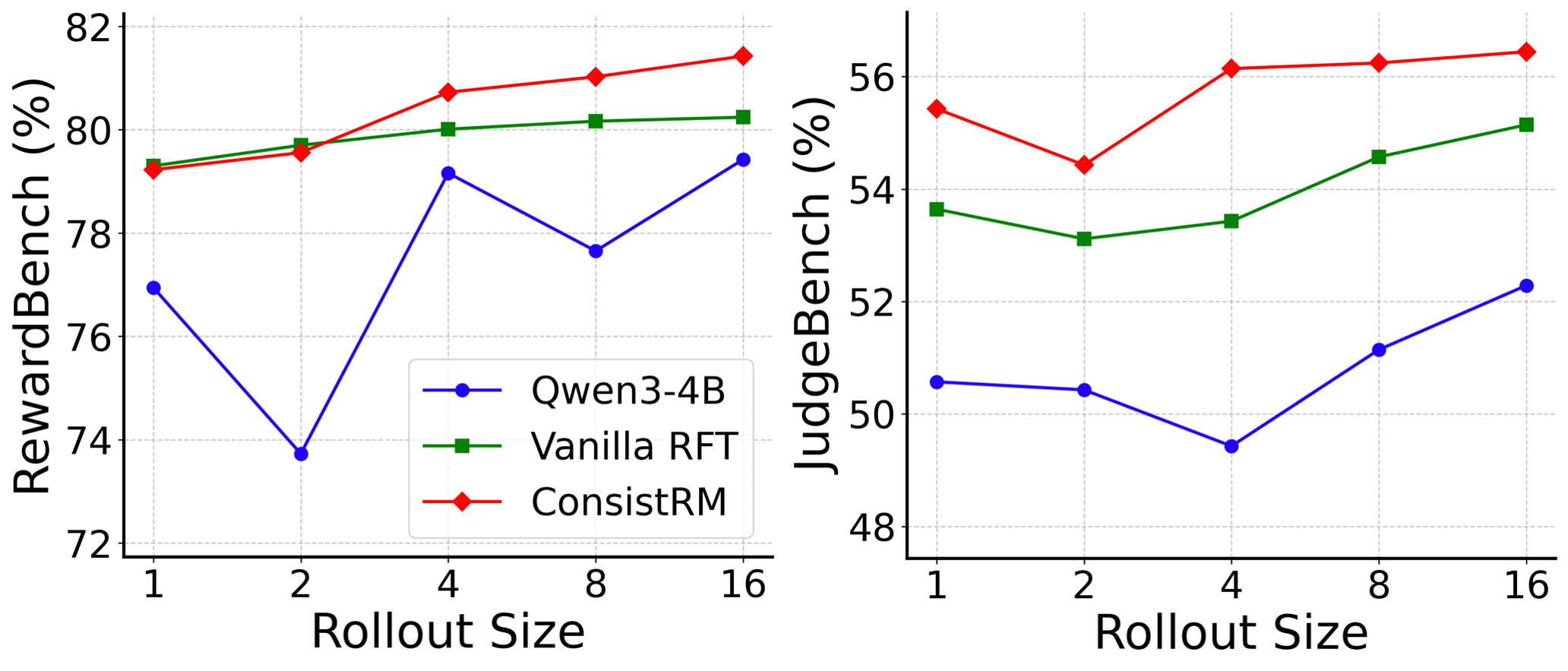}}
	\caption{Comparison of the stability of different GRMs with different rollout sizes. 
}
	\label{ConsistRM-res-rollout}
\end{figure}

\paragraph{Stability to Multiple Votes}
We further analyze the output stability of ConsistRM under high-temperature decoding, where stochasticity in generation can lead to increased variance in model outputs. To mitigate this issue, multi-round voting~\cite{wang2023selfconsistency} improves answer consistency by aggregating outputs from multiple reasoning paths sampled at high temperature.
We evaluate three models: the Qwen3-4B model, vanilla RFT, and our ConsistRM on RewardBench and JudgeBench, varying the number of rollout sizes (1, 2, 4, 8, 16). 
As shown in Figure~\ref{ConsistRM-res-rollout}, when applied to the base model, multi-round voting is observed to yield unstable gains on generative reward tasks due to output inconsistency, whereas vanilla RFT provides stable but limited improvements.
In contrast, ConsistRM achieves consistent performance gains as the number of paths increases, eventually matching the performance of an 8B model while retaining the 4B parameter scale. This improvement can be attributed to ConsistRM's ability to effectively supervise the critique process, thereby enhancing response stability.

\begin{figure}[t]
	\centering
\scalebox{0.4}{
	\includegraphics[]{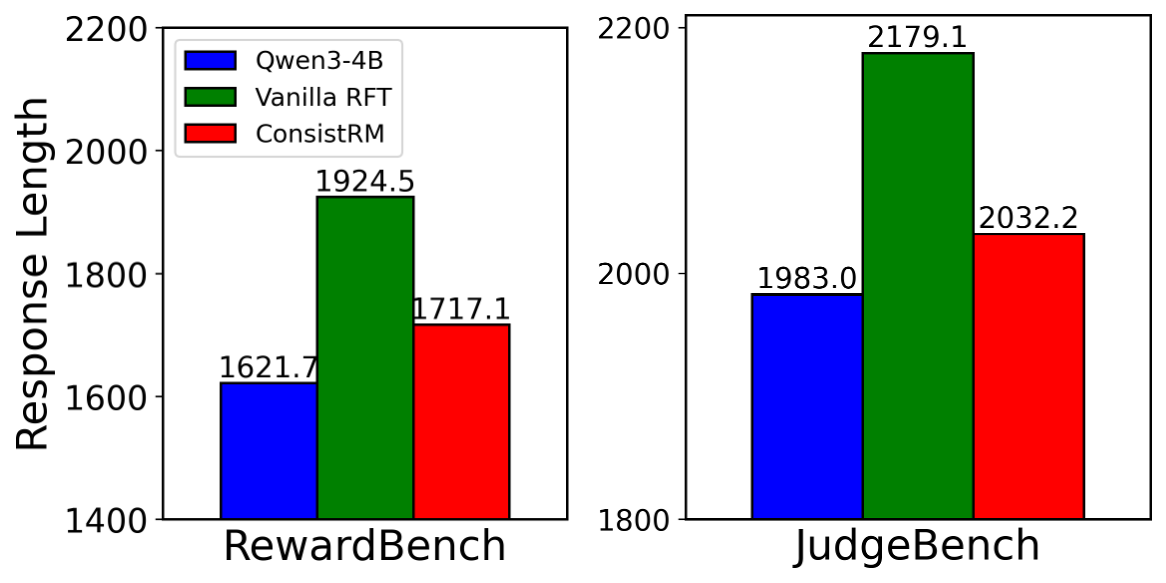}}
	\caption{Changing trends of the average response length with different GRMs. ConsistRM achieves the balance between performance and efficiency. }
	\label{ConsistRM-res-length}
\end{figure}
\subsection{Less tokens, More power}
\paragraph{The efficiency of ConsistRM}
\label{Ablation Study3 of ConsistRM}


ConsistRM is a pairwise preference reward model designed to improve both task performance and response efficiency—the latter measured by output length. To evaluate efficiency gains, we compare the average response lengths of Qwen3-4B, vanilla RFT, and ConsistRM on RewardBench and JudgeBench. As shown in Figure \ref{ConsistRM-res-length}, RFT produces significantly longer responses than Qwen3-4B, likely because  RFT relies solely on outcome reward without explicit length constraints. In contrast, ConsistRM effectively controls output length while maintaining competitive performance. For example, on RewardBench, it reduces the average response length from 1,924.5 to 1,717.1. 
This reduction is primarily driven by consistency-aware critique rewards, which improve task performance while suppressing excessive verbosity in model outputs.

\section*{Conclusion}
This paper proposes ConsistRM, a novel self-training strategy for GRMs that leverages the consistency awareness of LLMs to generate supervisory signals without relying on external human feedback.
ConsistRM consists of two key modules: the Consistency-Aware Label Reward module and the Consistency-Aware Critique Reward module. 
The former enables self-training by explicitly modeling the model’s internal preference consistency across multiple levels, while the latter derives more stable evaluation criteria to provide richer reward signals.
Extensive experiments on multiple benchmarks and models of varying sizes demonstrate the effectiveness and robustness of the proposed ConsistRM. 
Furthermore, our analysis shows that the approach significantly improves the consistency and efficiency of answer generation while mitigating positional bias.

\section*{Limitations}
Although ConsistRM demonstrates significant performance improvements, it has several limitations. First, semantic consistency is currently evaluated only at the overall critique level, which may overlook fine-grained alignment between text segments. Future work could address this by segmenting critiques into smaller semantic units and providing reward supervision at a more granular level. This direction aligns with research on process reward modeling in reinforcement learning, which aims to provide verifiable rewards and remains an open avenue for future exploration.

\section*{Ethics Statement}
This work follows the ACL Ethics Policy. Our findings are based on publicly available datasets for reproducibility purposes. We acknowledge that LLMs may exhibit inherent societal biases and are prone to hallucinations. Therefore, if someone finds our work interesting and would like to use it in a specific environment, we strongly suggest the user conduct safety and bias evaluations to mitigate potential risks.


\begin{thebibliography}{43}
\providecommand{\natexlab}[1]{#1}

\bibitem[{Azar et~al.(2024)Azar, Guo, Piot, Munos, Rowland, Valko, and Calandriello}]{azar2024general}
Mohammad~Gheshlaghi Azar, Zhaohan~Daniel Guo, Bilal Piot, Remi Munos, Mark Rowland, Michal Valko, and Daniele Calandriello. 2024.
\newblock A general theoretical paradigm to understand learning from human preferences.
\newblock In \emph{International Conference on Artificial Intelligence and Statistics}, pages 4447--4455. PMLR.

\bibitem[{Bai et~al.(2022)Bai, Jones, Ndousse, Askell, Chen, DasSarma, Drain, Fort, Ganguli, Henighan et~al.}]{bai2022training}
Yuntao Bai, Andy Jones, Kamal Ndousse, Amanda Askell, Anna Chen, Nova DasSarma, Dawn Drain, Stanislav Fort, Deep Ganguli, Tom Henighan, and 1 others. 2022.
\newblock Training a helpful and harmless assistant with reinforcement learning from human feedback.
\newblock \emph{arXiv preprint arXiv:2204.05862}.

\bibitem[{Chen et~al.(2025)Chen, Li, Wang, Jin, Qian, Wang, Wang, Zhang, Zhang, Zhang et~al.}]{chen2025rm}
Xiusi Chen, Gaotang Li, Ziqi Wang, Bowen Jin, Cheng Qian, Yu~Wang, Hongru Wang, Yu~Zhang, Denghui Zhang, Tong Zhang, and 1 others. 2025.
\newblock Rm-r1: Reward modeling as reasoning.
\newblock \emph{arXiv preprint arXiv:2505.02387}.

\bibitem[{Dong et~al.(2024)Dong, Xiong, Pang, Wang, Zhao, Zhou, Jiang, Sahoo, Xiong, and Zhang}]{dong2024rlhf}
Hanze Dong, Wei Xiong, Bo~Pang, Haoxiang Wang, Han Zhao, Yingbo Zhou, Nan Jiang, Doyen Sahoo, Caiming Xiong, and Tong Zhang. 2024.
\newblock Rlhf workflow: From reward modeling to online rlhf.
\newblock \emph{arXiv preprint arXiv:2405.07863}.

\bibitem[{Frick et~al.(2025)Frick, Li, Chen, Chiang, Angelopoulos, Jiao, Zhu, Gonzalez, and Stoica}]{frick2025how}
Evan Frick, Tianle Li, Connor Chen, Wei-Lin Chiang, Anastasios~Nikolas Angelopoulos, Jiantao Jiao, Banghua Zhu, Joseph~E. Gonzalez, and Ion Stoica. 2025.
\newblock \href {https://openreview.net/forum?id=cbttLtO94Q} {{How to Evaluate Reward Models for {RLHF}}}.
\newblock In \emph{The Thirteenth International Conference on Learning Representations}.

\bibitem[{Fu et~al.(2025)Fu, Wang, Tian, and Zhao}]{fu2025deep}
Yichao Fu, Xuewei Wang, Yuandong Tian, and Jiawei Zhao. 2025.
\newblock Deep think with confidence.
\newblock \emph{arXiv preprint arXiv:2508.15260}.

\bibitem[{Gao et~al.(2023)Gao, Schulman, and Hilton}]{gao2023scaling}
Leo Gao, John Schulman, and Jacob Hilton. 2023.
\newblock Scaling laws for reward model overoptimization.
\newblock In \emph{International Conference on Machine Learning}, pages 10835--10866. PMLR.

\bibitem[{Guo et~al.(2025{\natexlab{a}})Guo, Yang, Zhang, Song, Zhang, Xu, Zhu, Ma, Wang, Bi et~al.}]{guo2025deepseek}
Daya Guo, Dejian Yang, Haowei Zhang, Junxiao Song, Ruoyu Zhang, Runxin Xu, Qihao Zhu, Shirong Ma, Peiyi Wang, Xiao Bi, and 1 others. 2025{\natexlab{a}}.
\newblock Deepseek-r1: Incentivizing reasoning capability in llms via reinforcement learning.
\newblock \emph{arXiv preprint arXiv:2501.12948}.

\bibitem[{Guo et~al.(2025{\natexlab{b}})Guo, Chi, Dong, Dong, Wu, Huang, and Wei}]{guo2025reward}
Jiaxin Guo, Zewen Chi, Li~Dong, Qingxiu Dong, Xun Wu, Shaohan Huang, and Furu Wei. 2025{\natexlab{b}}.
\newblock Reward reasoning model.
\newblock \emph{arXiv preprint arXiv:2505.14674}.

\bibitem[{Lai et~al.(2025)Lai, Liu, Gao, Cheng, Qi, Xu, Yao, Zhang, Du, Hou et~al.}]{lai2025survey}
Hanyu Lai, Xiao Liu, Junjie Gao, Jiale Cheng, Zehan Qi, Yifan Xu, Shuntian Yao, Dan Zhang, Jinhua Du, Zhenyu Hou, and 1 others. 2025.
\newblock A survey of post-training scaling in large language models.
\newblock In \emph{Proceedings of the 63rd Annual Meeting of the Association for Computational Linguistics (Volume 1: Long Papers)}, pages 2771--2791.

\bibitem[{Laidlaw et~al.(2024)Laidlaw, Singhal, and Dragan}]{laidlaw2024correlated}
Cassidy Laidlaw, Shivam Singhal, and Anca Dragan. 2024.
\newblock Correlated proxies: A new definition and improved mitigation for reward hacking.
\newblock \emph{arXiv preprint arXiv:2403.03185}.

\bibitem[{Lambert(2025)}]{lambert2025reinforcement}
Nathan Lambert. 2025.
\newblock Reinforcement learning from human feedback.
\newblock \emph{arXiv preprint arXiv:2504.12501}.

\bibitem[{Lambert et~al.(2025)Lambert, Pyatkin, Morrison, Miranda, Lin, Chandu, Dziri, Kumar, Zick, Choi, Smith, and Hajishirzi}]{lambert2024rewardbench}
Nathan Lambert, Valentina Pyatkin, Jacob Morrison, LJ~Miranda, Bill~Yuchen Lin, Khyathi Chandu, Nouha Dziri, Sachin Kumar, Tom Zick, Yejin Choi, Noah~A. Smith, and Hannaneh Hajishirzi. 2025.
\newblock \href {https://doi.org/10.18653/v1/2025.findings-naacl.96} {{Rewardbench: Evaluating Reward Models for Language Modeling}}.
\newblock In \emph{Findings of the Association for Computational Linguistics: NAACL 2025}, pages 1755--1797, Albuquerque, New Mexico. Association for Computational Linguistics.

\bibitem[{Li et~al.(2025{\natexlab{a}})Li, Zhang, Yu, Saha, Khashabi, Weston, Lanchantin, and Wang}]{li2025jointly}
Tianjian Li, Yiming Zhang, Ping Yu, Swarnadeep Saha, Daniel Khashabi, Jason Weston, Jack Lanchantin, and Tianlu Wang. 2025{\natexlab{a}}.
\newblock Jointly reinforcing diversity and quality in language model generations.
\newblock \emph{arXiv preprint arXiv:2509.02534}.

\bibitem[{Li et~al.(2025{\natexlab{b}})Li, Xu, Yu, Zhang, Chen, Ling, Chao, Yuan, and Zhou}]{li2025generalist}
Yi-Chen Li, Tian Xu, Yang Yu, Xuqin Zhang, Xiong-Hui Chen, Zhongxiang Ling, Ningjing Chao, Lei Yuan, and Zhi-Hua Zhou. 2025{\natexlab{b}}.
\newblock Generalist reward models: Found inside large language models.
\newblock \emph{arXiv preprint arXiv:2506.23235}.

\bibitem[{Liu et~al.(2025{\natexlab{a}})Liu, Zeng, Xiao, He, Liu, Wang, Yan, Shen, Zhang, Xu, Liu, and Zhou}]{liu2025skywork}
Chris~Yuhao Liu, Liang Zeng, Yuzhen Xiao, Jujie He, Jiacai Liu, Chaojie Wang, Rui Yan, Wei Shen, Fuxiang Zhang, Jiacheng Xu, Yang Liu, and Yahui Zhou. 2025{\natexlab{a}}.
\newblock Skywork-reward-v2: Scaling preference data curation via human-ai synergy.
\newblock \emph{arXiv preprint arXiv:2507.01352}.

\bibitem[{Liu et~al.(2025{\natexlab{b}})Liu, He, Lin, Yang, Shen, and Liu}]{liu2025ettrl}
Jia Liu, ChangYi He, YingQiao Lin, MingMin Yang, FeiYang Shen, and ShaoGuo Liu. 2025{\natexlab{b}}.
\newblock Ettrl: Balancing exploration and exploitation in llm test-time reinforcement learning via entropy mechanism.
\newblock \emph{arXiv preprint arXiv:2508.11356}.

\bibitem[{Liu et~al.(2024)Liu, Nguyen, Shang, Ding, Li, Yu, Kumar, and Wang}]{liu2024learning}
Jiawei Liu, Thanh Nguyen, Mingyue Shang, Hantian Ding, Xiaopeng Li, Yu~Yu, Varun Kumar, and Zijian Wang. 2024.
\newblock Learning code preference via synthetic evolution.
\newblock \emph{arXiv preprint arXiv:2410.03837}.

\bibitem[{Liu et~al.(2025{\natexlab{c}})Liu, Yao, Min, Cao, Hou, and Li}]{liu2025rmbench}
Yantao Liu, Zijun Yao, Rui Min, Yixin Cao, Lei Hou, and Juanzi Li. 2025{\natexlab{c}}.
\newblock \href {https://openreview.net/forum?id=QEHrmQPBdd} {{{RM}-Bench: Benchmarking Reward Models of Language Models with Subtlety and Style}}.
\newblock In \emph{The Thirteenth International Conference on Learning Representations}.

\bibitem[{Liu et~al.(2025{\natexlab{d}})Liu, Wang, Xu, Ma, Ruan, Li, Liu, and Wu}]{liu2025inference}
Zijun Liu, Peiyi Wang, Runxin Xu, Shirong Ma, Chong Ruan, Peng Li, Yang Liu, and Yu~Wu. 2025{\natexlab{d}}.
\newblock Inference-time scaling for generalist reward modeling.
\newblock \emph{arXiv preprint arXiv:2504.02495}.

\bibitem[{Mahan et~al.(2024)Mahan, Van~Phung, Rafailov, Blagden, Lile, Castricato, Fr{\"a}nken, Finn, and Albalak}]{mahan2024generative}
Dakota Mahan, Duy Van~Phung, Rafael Rafailov, Chase Blagden, Nathan Lile, Louis Castricato, Jan-Philipp Fr{\"a}nken, Chelsea Finn, and Alon Albalak. 2024.
\newblock Generative reward models.
\newblock \emph{arXiv preprint arXiv:2410.12832}.

\bibitem[{Ouyang et~al.(2022)Ouyang, Wu, Jiang, Almeida, Wainwright, Mishkin, Zhang, Agarwal, Slama, Ray et~al.}]{ouyang2022training}
Long Ouyang, Jeffrey Wu, Xu~Jiang, Diogo Almeida, Carroll Wainwright, Pamela Mishkin, Chong Zhang, Sandhini Agarwal, Katarina Slama, Alex Ray, and 1 others. 2022.
\newblock Training language models to follow instructions with human feedback.
\newblock \emph{Advances in neural information processing systems}, 35:27730--27744.

\bibitem[{Rafailov et~al.(2023)Rafailov, Sharma, Mitchell, Manning, Ermon, and Finn}]{rafailov2023direct}
Rafael Rafailov, Archit Sharma, Eric Mitchell, Christopher~D Manning, Stefano Ermon, and Chelsea Finn. 2023.
\newblock Direct preference optimization: Your language model is secretly a reward model.
\newblock \emph{Advances in neural information processing systems}, 36:53728--53741.

\bibitem[{Saha et~al.(2025)Saha, Li, Ghazvininejad, Weston, and Wang}]{saha2025learning}
Swarnadeep Saha, Xian Li, Marjan Ghazvininejad, Jason Weston, and Tianlu Wang. 2025.
\newblock Learning to plan \& reason for evaluation with thinking-llm-as-a-judge.
\newblock \emph{arXiv preprint arXiv:2501.18099}.

\bibitem[{Shafayat et~al.(2025)Shafayat, Tajwar, Salakhutdinov, Schneider, and Zanette}]{shafayat2025can}
Sheikh Shafayat, Fahim Tajwar, Ruslan Salakhutdinov, Jeff Schneider, and Andrea Zanette. 2025.
\newblock Can large reasoning models self-train?
\newblock \emph{arXiv preprint arXiv:2505.21444}.

\bibitem[{Shao et~al.(2024)Shao, Wang, Zhu, Xu, Song, Bi, Zhang, Zhang, Li, Wu et~al.}]{shao2024deepseekmath}
Zhihong Shao, Peiyi Wang, Qihao Zhu, Runxin Xu, Junxiao Song, Xiao Bi, Haowei Zhang, Mingchuan Zhang, YK~Li, Yang Wu, and 1 others. 2024.
\newblock Deepseekmath: Pushing the limits of mathematical reasoning in open language models.
\newblock \emph{arXiv preprint arXiv:2402.03300}.

\bibitem[{Tan et~al.(2025)Tan, Zhuang, Montgomery, Tang, Cuadron, Wang, Popa, and Stoica}]{tan2025judgebench}
Sijun Tan, Siyuan Zhuang, Kyle Montgomery, William~Yuan Tang, Alejandro Cuadron, Chenguang Wang, Raluca Popa, and Ion Stoica. 2025.
\newblock \href {https://openreview.net/forum?id=G0dksFayVq} {{JudgeBench: A Benchmark for Evaluating {LLM}-Based Judges}}.
\newblock In \emph{The Thirteenth International Conference on Learning Representations}.

\bibitem[{van Niekerk et~al.(2025)van Niekerk, Vukovic, Ruppik, Lin, and Ga{\v{s}}i{\'c}}]{van2025post}
Carel van Niekerk, Renato Vukovic, Benjamin~Matthias Ruppik, Hsien-chin Lin, and Milica Ga{\v{s}}i{\'c}. 2025.
\newblock Post-training large language models via reinforcement learning from self-feedback.
\newblock \emph{arXiv preprint arXiv:2507.21931}.

\bibitem[{Wang et~al.(2025{\natexlab{a}})Wang, Wang, Chen, and Huang}]{wang2025beyond}
Weiqin Wang, Yile Wang, Kehao Chen, and Hui Huang. 2025{\natexlab{a}}.
\newblock Beyond majority voting: Towards fine-grained and more reliable reward signal for test-time reinforcement learning.
\newblock \emph{arXiv preprint arXiv:2512.15146}.

\bibitem[{Wang et~al.(2023)Wang, Wei, Schuurmans, Le, Chi, Narang, Chowdhery, and Zhou}]{wang2023selfconsistency}
Xuezhi Wang, Jason Wei, Dale Schuurmans, Quoc~V Le, Ed~H. Chi, Sharan Narang, Aakanksha Chowdhery, and Denny Zhou. 2023.
\newblock \href {https://openreview.net/forum?id=1PL1NIMMrw} {Self-consistency improves chain of thought reasoning in language models}.
\newblock In \emph{The Eleventh International Conference on Learning Representations}.

\bibitem[{Wang et~al.(2025{\natexlab{b}})Wang, Zeng, Delalleau, Shin, Soares, Bukharin, Evans, Dong, and Kuchaiev}]{wang2025helpsteer3}
Zhilin Wang, Jiaqi Zeng, Olivier Delalleau, Hoo-Chang Shin, Felipe Soares, Alexander Bukharin, Ellie Evans, Yi~Dong, and Oleksii Kuchaiev. 2025{\natexlab{b}}.
\newblock Helpsteer3-preference: Open human-annotated preference data across diverse tasks and languages.
\newblock \emph{arXiv preprint arXiv:2505.11475}.

\bibitem[{Wang et~al.(2025{\natexlab{c}})Wang, Xu, Wang, Ye, Li, Wang, Tan, Wang, Feng, Chen et~al.}]{wang2025survey}
Zihan Wang, Xingle Xu, Hao Wang, Yiwen Ye, Yuchen Li, Linhao Wang, Hongze Tan, Peidong Wang, Shi Feng, Guoqing Chen, and 1 others. 2025{\natexlab{c}}.
\newblock A survey on entropy mechanism in large reasoning models.
\newblock \emph{Authorea Preprints}.

\bibitem[{Whitehouse et~al.(2025)Whitehouse, Wang, Yu, Li, Weston, Kulikov, and Saha}]{whitehouse2025j1}
Chenxi Whitehouse, Tianlu Wang, Ping Yu, Xian Li, Jason Weston, Ilia Kulikov, and Swarnadeep Saha. 2025.
\newblock J1: Incentivizing thinking in llm-as-a-judge via reinforcement learning.
\newblock \emph{arXiv preprint arXiv:2505.10320}.

\bibitem[{Yang et~al.(2025)Yang, Li, Yang, Zhang, Hui, Zheng, Yu, Gao, Huang, Lv et~al.}]{yang2025qwen3}
An~Yang, Anfeng Li, Baosong Yang, Beichen Zhang, Binyuan Hui, Bo~Zheng, Bowen Yu, Chang Gao, Chengen Huang, Chenxu Lv, and 1 others. 2025.
\newblock Qwen3 technical report.
\newblock \emph{arXiv preprint arXiv:2505.09388}.

\bibitem[{Ye et~al.(2025)Ye, Greenlee, Bartolo, Blunsom, Campos, and Gall{\'e}}]{ye2025improving}
Zihuiwen Ye, Fraser~David Greenlee, Max Bartolo, Phil Blunsom, Jon~Ander Campos, and Matthias Gall{\'e}. 2025.
\newblock Improving reward models with synthetic critiques.
\newblock In \emph{Findings of the Association for Computational Linguistics: NAACL 2025}, pages 4506--4520.

\bibitem[{Zhang et~al.(2025{\natexlab{a}})Zhang, Yao, Liu, Wang, Lai, Ye, Song, and Tao}]{zhang2025consistent}
Kongcheng Zhang, Qi~Yao, Shunyu Liu, Yingjie Wang, Baisheng Lai, Jieping Ye, Mingli Song, and Dacheng Tao. 2025{\natexlab{a}}.
\newblock Consistent paths lead to truth: Self-rewarding reinforcement learning for llm reasoning.
\newblock \emph{arXiv preprint arXiv:2506.08745}.

\bibitem[{Zhang et~al.(2025{\natexlab{b}})Zhang, Li, Long, Zhang, Lin, Yang, Xie, Yang, Liu, Lin, Huang, and Zhou}]{qwen3embedding}
Yanzhao Zhang, Mingxin Li, Dingkun Long, Xin Zhang, Huan Lin, Baosong Yang, Pengjun Xie, An~Yang, Dayiheng Liu, Junyang Lin, Fei Huang, and Jingren Zhou. 2025{\natexlab{b}}.
\newblock Qwen3 embedding: Advancing text embedding and reranking through foundation models.
\newblock \emph{arXiv preprint arXiv:2506.05176}.

\bibitem[{Zhang et~al.(2025{\natexlab{c}})Zhang, Zhang, Guan, Cheng, Duan, Wang, Wang, Zheng, and He}]{zhang2025no}
Yanzhi Zhang, Zhaoxi Zhang, Haoxiang Guan, Yilin Cheng, Yitong Duan, Chen Wang, Yue Wang, Shuxin Zheng, and Jiyan He. 2025{\natexlab{c}}.
\newblock No free lunch: Rethinking internal feedback for llm reasoning.
\newblock \emph{arXiv preprint arXiv:2506.17219}.

\bibitem[{Zhang et~al.(2025{\natexlab{d}})Zhang, Zhu, Ge, Zhao, Zhou, Li, Feng, Yao, and Han}]{zhang2025co}
Zizhuo Zhang, Jianing Zhu, Xinmu Ge, Zihua Zhao, Zhanke Zhou, Xuan Li, Xiao Feng, Jiangchao Yao, and Bo~Han. 2025{\natexlab{d}}.
\newblock Co-rewarding: Stable self-supervised rl for eliciting reasoning in large language models.
\newblock \emph{arXiv preprint arXiv:2508.00410}.

\bibitem[{Zhao et~al.(2025)Zhao, Kang, Feng, Levine, and Song}]{zhao2505learning}
Xuandong Zhao, Zhewei Kang, Aosong Feng, Sergey Levine, and Dawn Song. 2025.
\newblock Learning to reason without external rewards.
\newblock \emph{URL https://arxiv. org/abs/2505.19590}, 2.

\bibitem[{Zhou et~al.(2024)Zhou, Zheng, Wang, Xi, Dou, Bao, Shen, Xiong, Fan, Mou et~al.}]{zhou2024rmb}
Enyu Zhou, Guodong Zheng, Binghai Wang, Zhiheng Xi, Shihan Dou, Rong Bao, Wei Shen, Limao Xiong, Jessica Fan, Yurong Mou, and 1 others. 2024.
\newblock Rmb: Comprehensively benchmarking reward models in llm alignment.
\newblock \emph{arXiv preprint arXiv:2410.09893}.

\bibitem[{Zhou et~al.(2025)Zhou, Liang, Liu, Yu, Panaganti, Song, Yu, Zhang, Mi, and Yu}]{zhou2025evolving}
Yujun Zhou, Zhenwen Liang, Haolin Liu, Wenhao Yu, Kishan Panaganti, Linfeng Song, Dian Yu, Xiangliang Zhang, Haitao Mi, and Dong Yu. 2025.
\newblock Evolving language models without labels: Majority drives selection, novelty promotes variation.
\newblock \emph{arXiv preprint arXiv:2509.15194}.

\bibitem[{Zuo et~al.(2025)Zuo, Zhang, Sheng, Qu, Cui, Zhu, Li, Zhang, Long, Hua et~al.}]{zuo2025ttrl}
Yuxin Zuo, Kaiyan Zhang, Li~Sheng, Shang Qu, Ganqu Cui, Xuekai Zhu, Haozhan Li, Yuchen Zhang, Xinwei Long, Ermo Hua, and 1 others. 2025.
\newblock Ttrl: Test-time reinforcement learning.
\newblock \emph{arXiv preprint arXiv:2504.16084}.

\end{thebibliography}

\clearpage
\appendix

\section{Training Configuration}

\subsection{Hyperparameter Settings}
\label{A1}
We implement our method using the Volcano Engine Reinforcement Learning for LLMs framework and conduct our experiments on two recent open-source base models: Qwen3-4B and Qwen3-8B. All models are optimized using the GRPO algorithm for 4 training epochs. We adopt a global batch size of 64 and a learning rate of 1e-6. The maximum generation length is set to 1024 tokens and the maximum prompt length is set to 4096 tokens. During training, we use a PPO mini-batch size of 32 and perform 8 rollouts per update step. The generation temperature is fixed at 1.0 and the top-p is set to 0.8 to encourage exploration. We additionally enable KL regularization to stabilize training, with a KL loss coefficient of 0.001.

\begin{table}[ht]
\centering
\small
\begin{tabular}{lc}
\toprule
\textbf{Hyperparameter} & \textbf{Value} \\
\midrule
Training Epochs & 4 \\
Global Batch Size & 64 \\
Learning Rate & 1e-6 \\
Maximum Prompt Length & 4096 tokens \\
Maximum Generation Length & 1024 tokens \\
PPO Mini-Batch Size & 32 \\
Rollouts per Update Step & 8 \\
Clip Ratio & 0.2 \\
Training Temperature & 1.0 \\
Top-p & 0.8 \\
Use KL Loss & True \\
KL Loss Coefficient & 0.001 \\
\bottomrule
\end{tabular}
\caption{Implementation Details of Our Method}
\label{tab:implement_details}
\end{table}

\subsection{Training Data Construction for ConsistRM}
\label{A2}

In this study, rather than utilizing the full HelpSteer3 training dataset, we selectively curated the training data by leveraging the model itself for data filtering. Specifically, we first identified and removed examples for which the model produced outputs identical to the ground truth across all 8 rollouts. The remaining subset of data was subsequently used to construct the final training dataset. The detailed statistics of the resulting training data are summarized in Table~\ref{tab:training-data}.

\begin{table}[h!]
\centering
\small
\begin{tabular}{lc}
\toprule
Model & Number of Examples \\
\midrule
Qwen3-4B & 13,236 \\
Qwen3-8B & 12,098 \\
Qwen3-14B & 10,340 \\
\bottomrule
\end{tabular}
\caption{Statistics of the final training dataset}
\label{tab:training-data}
\end{table}

\subsection{Computational Resources}
All experiments reported in this paper were conducted on a single server with 8×NVIDIA A800 GPUs, while Qwen3-14B training adopts two servers.

\section{More Analysis of ConsistRM}
\subsection{Effectiveness of CAAR}
\label{Effectiveness of CAAR}
To further demonstrate the effectiveness of CAAR, Table~\ref{tab:pseudo_label_accuracy} shows that our method effectively combines Online-State Consistency Preference and Memory-Driven Consistency Preference, leading to a steady improvement in the accuracy of consistency-aware pseudo-labels (from 80.25\% to 91.5\%) as training progresses. This trend demonstrates that the proposed mechanism can provide increasingly reliable and dynamically adjusted supervision signals, enabling more stable learning based on high-quality pseudo-labels in the later stages of training.

\begin{table}[htbp]
\centering
\small
\setlength{\tabcolsep}{3pt}
\renewcommand{\arraystretch}{1.1}
\begin{tabular}{lccccc}
\toprule
 & 0 steps & 80 steps & 160 steps & 240 steps & 320 steps \\
\midrule
Acc (\%) & 80.25 & 83.45 & 87.31 & 88.25 & 91.50 \\
\bottomrule
\end{tabular}
\caption{Accuracy of consistency-aware pseudo-labels over training steps.}
\label{tab:pseudo_label_accuracy}
\end{table}

\subsection{Generalization of ConsistRM}
To assess the generalization of ConsistRM, we train the model on the general-purpose HelpSteer3 and evaluate its performance on the code-specific preference benchmark, CodePrefBench~\cite{liu2024learning}. The results shown in Table~\ref{tab:codeprefbench} indicate that ConsistRM consistently outperforms the RFT baseline under training–test distribution mismatch. Despite being trained on a general-domain dataset, it achieves substantial gains on this code-oriented benchmark, highlighting its strong generalization under distribution shift. Moreover, ConsistRM consistently outperforms RFT across model scales when trained on HelpSteer3.
\begin{table}[htbp]
\centering
\small
\setlength{\tabcolsep}{5pt}
\renewcommand{\arraystretch}{1.1}
\begin{tabular}{lcc}
\toprule
Model & CodePrefBench & $\Delta$ \\
\midrule
Qwen3-4B + RFT & 62.1 & -- \\
Qwen3-4B + ConsistRM & 66.6 & \textcolor[RGB]{0,176,80}{\textbf{+4.5}} \\
\midrule
Qwen3-8B + RFT & 69.7 & -- \\
Qwen3-8B + ConsistRM & 72.1 & \textcolor[RGB]{0,176,80}{\textbf{+2.4}} \\
\bottomrule
\end{tabular}
\caption{Performance on CodePrefBench.}
\label{tab:codeprefbench}
\end{table}

\subsection{Robustness of ConsistRM}
\label{Robustness of ConsistRM}
To investigate the impact of incorrect pseudo-labels (i.e., false consistency), we analyze samples that conflict with the gold preference labels, as such inconsistencies naturally arise in consistency-aware pseudo-labeling. To assess their impact on model performance, we remove these inconsistent samples during training and retrain the model under identical settings. As illustrated in Table~\ref{tab:incorrect_pseudo_labels}, this removal results in only a marginal performance change (from 73.5 to 73.3), indicating that such samples provide limited but non-negligible learning signals.  This is because these erroneous samples constitute only a small fracion of the training data in later stages (approximately 9.5\%), thereby limiting their overall influence on model optimization. 
\begin{table}[htbp]
\centering
\small
\setlength{\tabcolsep}{3pt}
\renewcommand{\arraystretch}{1}
\begin{tabular}{lcccccc}
\toprule
System & \makecell{Reward\\Bench} & \makecell{PPE\\Pref} & \makecell{RM-\\Bench} & RMB & \makecell{Judge\\Bench} & Avg \\
\midrule
\textbf{ConsistRM} & 85.6 & 67.7 & 78.3 & 79.1 & 56.9 & 73.5 \\
\quad w/ removal & 86.1 & 67.9 & 76.2 & 79.2 & 57.2 & 73.3 \\
\bottomrule
\end{tabular}
\caption{Effect of eliminating incorrect pseudo-labels on model performance.}
\label{tab:incorrect_pseudo_labels}
\end{table}

\cleardoublepage
\onecolumn
\section{Algorithm of ConsistRM}
We present the detailed pseudo-code of ConsistRM in Algorithm~\ref{alg:consistrm}. The algorithm outlines the full consistency-aware self-training pipeline, including pseudo-label construction and reward computation.

\label{Algorithm}
\begin{algorithm}[H]
\caption{ConsistRM: Consistency-Aware Self-Training for Generative Reward Models}
\label{alg:consistrm}
\begin{algorithmic}[1]

\State \textbf{Input:} $x = (q, a_1, a_2)$, rollout count $K$, experience buffer $\mathcal{E}$, GRM $\pi_\theta$
\State \textbf{Output:} Updated $\mathcal{E}$, rewards $\{r_j^{(n)}\}$  

\For{$j = 1$ to $K$}
    \State $(c_j^{(n)}, y_j^{(n)}) \sim \pi_\theta(x)$
\EndFor

\State $s_{\text{online}}^{(n)} \gets \frac{1}{K} \sum_{j} y_j^{(n)}$

\If{$|\mathcal{E}| > 0$}
    \State $s_{\text{memory}}^{(n)} \gets \frac{1}{|\mathcal{E}|} \sum_{\hat{y} \in \mathcal{E}} \hat{y}$
\Else
    \State $s_{\text{memory}}^{(n)} \gets 0$
\EndIf

\State $\hat{y}^{(n)} \gets \text{sgn}(s_{\text{online}}^{(n)} + s_{\text{memory}}^{(n)})$
\State $\mathcal{E} \gets \mathcal{E} \cup \{\hat{y}^{(n)}\}$

\State Compute embeddings $e_j^{(n)}$ for each $c_j^{(n)}$
\State Compute similarity scores and obtain ranking $\text{rank}(c_j^{(n)})$

\For{$j = 1$ to $K$}

    \If{$c_j^{(n)}$ is invalid}
        \State $r_j^{(n)} \gets -5$
    
    \ElsIf{$\hat{y}^{(n)} = 0$}
        \State $r_j^{(n)} \gets 0$
    
    \Else
        \State $r_j^{(a,n)} \gets 
        \begin{cases}
        1 & \text{if } y_j^{(n)} = \hat{y}^{(n)} \\
        -1 & \text{if } y_j^{(n)} \neq \hat{y}^{(n)}
        \end{cases}$

        \State $r_j^{(c,n)} \gets 
        \begin{cases}
        0.1 & \text{if } y_j^{(n)} = \hat{y}^{(n)} \land \text{rank}(c_j^{(n)}) \le p \\
        0 & \text{otherwise}
        \end{cases}$

        \State $r_j^{(n)} \gets r_j^{(a,n)} + r_j^{(c,n)}$
    
    \EndIf

\EndFor

\State \textbf{return} $\mathcal{E}$, $\{r_j^{(n)}\}$  

\end{algorithmic}
\end{algorithm}

\cleardoublepage
\onecolumn
\section{Prompts and Model Completions}
This section presents sample prompts and the responses generated by the models.
\subsection{Prompt for Training}
For all experiments, we adopted standardized templates to regulate the output format, and explicitly required the model to first define the evaluation dimensions and elaborate on the analysis process of the two answers, before presenting a clearly delimited final answer.

\begin{tcolorbox}[
  colback=white,               
  colframe=black,         
  boxrule=0.5pt,               
  arc=6pt,                  
  auto outer arc,           
  title={\textbf{Prompt Template.}},
  coltitle=white,             
  colbacktitle=gray,       
  fonttitle=\bfseries   
]
Initially, consider factors such as practicality, relevance, accuracy, comprehensiveness, originality, and detail richness. Include 2 to 4 criteria for high-quality responses within the \textless Criterion \textgreater and \textless/Criterion \textgreater. Subsequently, based on the \textless Criterion\textgreater, provide a thorough evaluation of each response within the \textless Analysis\textgreater and \textless/Analysis\textgreater. Avoid any positional biases and ensure that the sequence in which the responses were presented does not affect your judgement. The length of the responses should not sway your evaluation. Lastly, after completing your evaluation, deliver your final judgement strictly adhering to this format: "Response 1 is better than Response 2" OR "Response 2 is better than Response 1" within the \textless Result\textgreater and \textless/Result\textgreater.

\textbf{Output Format Requirements} \\
The special placeholders must be included. \\
\textless Criterion\textgreater \\
2 to 4 criteria for high-quality responses \\
\textless/Criterion\textgreater \\
\textless Analysis\textgreater \\
Response 1:\\
Evaluation of Response 1 based on the given Criterion \\
Response 2:\\
Evaluation of Response 2 based on the given Criterion \\
\textless/Analysis\textgreater \\
\textless Result\textgreater \\
Based on the output results from Criteria and Analysis, only print the following: "Response 1 is better than Response 2" OR "Response 2 is better than Response 1". \\
\textless/Result\textgreater

\textbf{Conversation Context \& Query}\\

\textbf{Responses to be Compared}\\

[The Begin of Response 1] \\

[The End of Response 1] \\

[The Begin of Response 2] \\

[The End of Response 2]
\end{tcolorbox}
\cleardoublepage
\subsection{Example from RewardBench Generation}
\begin{tcolorbox}[
  colback=white,                
  colframe=black,          
  boxrule=0.5pt,          
  arc=6pt,              
  auto outer arc,   
  title={\textbf{Example of RewardBench.}}, 
  coltitle=white,       
  colbacktitle=gray,        
  fonttitle=\bfseries      
]
\textbf{Conversation Context \& Query}\\
does the NHI use the top bottom or bottom up approach

\textbf{Response 1:}\\
The National Health Insurance (NHI) is a system that aims to provide universal healthcare coverage to all citizens of a country. The approach used by NHI can vary depending on the specific implementation and the country in question. However, in general, NHI systems tend to use a combination of both top-down and bottom-up approaches
A top-down approach in healthcare typically involves centralized decision-making and planning, with policies and guidelines set at the national level and implemented at the local level. This approach can ensure consistency, standardization, and equity in healthcare services across the country. In the context of NHI, a top-down approach might involve the government setting national health priorities, determining the benefits package, and establishing payment rates for healthcare providers.
On the other hand, a bottom-up approach in healthcare emphasizes local decision-making, community engagement, and the involvement of healthcare providers and patients in the planning and implementation of healthcare services. This approach can help ensure that healthcare services are responsive to local needs and preferences and can promote greater ownership and accountability among healthcare providers and communities. In the context of NHI, a bottom-up approach might involve local healthcare providers and communities participating in the design and implementation of healthcare services, with the national government providing support and resources.
In practice, many NHI systems use a combination of both top-down and bottom-up approaches. For example, the national government might set broad policy guidelines and provide funding, while local healthcare providers and communities have the flexibility to tailor healthcare services to meet local needs and preferences. 
In summary, the NHI can use both top-down and bottom-up approaches, depending on the specific implementation and the country in question. A combination of both approaches can help ensure that healthcare services are equitable, responsive to local needs, and efficient.

\textbf{Response 2:}\\
The National Health Insurance (NHI) system in Taiwan uses a bottom-up approach. This means that the NHI is funded primarily through premiums collected from the population, which are then used to pay for healthcare services. The system is designed to provide universal coverage and equal access to healthcare services for all citizens, with the government subsidizing the premiums for those who cannot afford them.
The bottom-up approach in the context of the NHI involves the following key features:
1. **Premium Collection**: The NHI is funded through premiums that are collected from the insured population. These premiums are based on income levels, with higher-income individuals paying higher premiums.
2. **Pooling of Funds**: The collected premiums are pooled together to create a large fund that is used to pay for healthcare services. This pooling of funds helps to spread the financial risk across the entire population, ensuring that everyone has access to healthcare when needed.
3. **Service Provision**: Healthcare providers, such as hospitals and clinics, are reimbursed by the NHI for the services they provide to insured individuals. The reimbursement rates are set by the government and are based on the type of service provided.
4. **Government Subsidies**: The government provides subsidies to help cover the cost of premiums for low-income individuals and families, ensuring that everyone has access to healthcare services regardless of their ability to pay.
In summary, the NHI in Taiwan uses a bottom-up approach, with premiums collected from the population being used to fund healthcare services. The system is designed to provide universal coverage and equal access to healthcare services for all citizens, with the government subsidizing the premiums for those who cannot afford them.

\textbf{Ground Truth:}\\
Response 1 is better than Response 2
\end{tcolorbox}
\cleardoublepage

\begin{tcolorbox}[
  colback=white,                
  colframe=black,          
  boxrule=0.5pt,          
  arc=6pt,              
  auto outer arc,   
  title={\textbf{Generation of Base GRM.}}, 
  coltitle=white,       
  colbacktitle=gray,        
  fonttitle=\bfseries      
]
\textless Criterion\textgreater \\
1. Relevance and Correctness: The response should accurately answer the user’s question and provide information that is factually correct regarding the National Health Insurance (NHI) and its top-down or bottom-up approach.\\
2. Depth and Coverage: The response should address all important aspects of the NHI system and its operational approach, offering a thorough and well-rounded explanation.\\
3. Organization and Readability: The response should be clearly structured, with ideas presented in a logical sequence and easy for readers to follow.\\
4. Detail and Exemplification: The response should include concrete details or examples where relevant, particularly when discussing specific implementations, such as the NHI in Taiwan.\\
\textless /Criterion\textgreater \\

\textless Analysis\textgreater \\
Response 1: \\ 
Response 1 provides a broad overview of NHI systems and correctly explains the concepts of both top-down and bottom-up approaches in healthcare. It is accurate in describing the general mechanisms and benefits of each approach, and it is well-organized and clear. However, the response remains largely abstract and does not focus on any specific country or concrete implementation. As a result, while it demonstrates strong general knowledge, it lacks concrete details that would help a user understand how the NHI operates in practice, making it less actionable for someone seeking a real-world example.\\

Response 2: \\
Response 2 focuses specifically on the NHI in Taiwan and provides a detailed explanation of how the bottom-up approach is implemented in practice. It accurately describes key mechanisms, including premium collection, pooling of funds, service reimbursement, and government subsidies, linking each to the bottom-up operational model. The response is structured clearly, presenting information in a logical sequence, and offers concrete examples that enhance understanding. While it does not discuss the top-down approach, the specificity, practical relevance, and clarity make it more directly useful and easier to apply to a real-world context compared with Response 1.\\
\textless /Analysis\textgreater \\

\textless Result\textgreater \textcolor{red}{Response 2 is better than Response 1} \textless /Result\textgreater
\end{tcolorbox}

\begin{tcolorbox}[
  colback=white,                
  colframe=black,          
  boxrule=0.5pt,          
  arc=6pt,              
  auto outer arc,   
  title={\textbf{Generation After CosistRM Training.}}, 
  coltitle=white,       
  colbacktitle=gray,        
  fonttitle=\bfseries      
]
\textless Criterion\textgreater \\
1. **Accuracy and Relevance**: The response must correctly address the user's question and provide accurate information relevant to the National Health Insurance (NHI) and the top-bottom or bottom-up approach.\\
2. **Comprehensiveness**: The response should cover all relevant aspects of the NHI system and its approach, providing a thorough explanation.\\
3. **Clarity and Structure**: The response must be clearly organized, with logical flow and easy to understand.\\
4. **Specificity**: The response should provide specific examples or details where applicable, especially when discussing specific implementations like the NHI in Taiwan.\\
\textless /Criterion\textgreater \\

\textless Analysis\textgreater \\
Response 1:\\
Response 1 provides a comprehensive overview of the NHI system and explains both top-down and bottom-up approaches in the context of healthcare. It correctly identifies that NHI systems can use a combination of both approaches, and it offers clear explanations of each. The response is accurate, well-structured, and relevant to the user's query. However, it is somewhat general and does not specify any particular country's NHI system, which could have been more informative given the nature of the question.\\

Response 2:\\
Response 2 is more specific, as it focuses on the NHI in Taiwan and clearly explains how it uses a bottom-up approach. It provides detailed features of the system, including premium collection, pooling of funds, service provision, government subsidies, and universal coverage. The response is accurate, relevant, and clearly structured. However, it assumes that the user is referring to the NHI in Taiwan without explicitly stating this, which may limit its applicability to other NHI systems around the world. It also does not mention the top-down approach at all, which could be seen as a limitation in addressing the broader question. \\
\textless /Analysis\textgreater \\

\textless Result\textgreater \textcolor{green}{Response 1 is better than Response 2} \textless /Result\textgreater
\end{tcolorbox}


\end{document}